\documentclass[journal]{IEEEtran}
\usepackage{cite}
\usepackage{amsmath,amssymb,amsfonts}
\usepackage{algorithmic}
\usepackage{graphicx,color}
\usepackage{array}
\usepackage{amsmath,amssymb,amsfonts,graphicx,url,subcaption,float,booktabs,hyperref, textcomp}
\usepackage{url}
\def\BibTeX{{\rm B\kern-.05em{\sc i\kern-.025em b}\kern-.08em
    T\kern-.1667em\lower.7ex\hbox{E}\kern-.125emX}}
\AtBeginDocument{\definecolor{ojcolor}{cmyk}{0.93,0.59,0.15,0.02}}

\begin{document}


\title{SemOD: Semantic Enabled Object Detection Network under Various Weather Conditions}

\author{Aiyinsi Zuo\thanks{Department of Electrical and Computer Engineering, University of Rochester, Rochester, NY 14627 USA}
\and Zhaoliang Zheng\thanks{Department of Electrical and Computer Engineering, University of California, Los Angeles, CA 90024 USA. Corresponding author: zhz03@g.ucla.edu.}
}

\maketitle





\maketitle

\begin{abstract}
In the field of autonomous driving, camera-based perception models are mostly trained on clear weather data. Models that focus on addressing specific weather challenges are unable to adapt to various weather changes and primarily prioritize their weather removal characteristics. Our study introduces a semantic-enabled network for object detection in diverse weather conditions. In our analysis, semantics information can enable the model to generate plausible content for missing areas, understand object boundaries, and preserve visual coherency and realism across both filled-in and existing portions of the image, which are conducive to image transformation and object recognition. Specific in implementation, our architecture consists of a Preprocessing Unit (PPU) and a Detection Unit (DTU), where the PPU utilizes a U-shaped net enriched by semantics to refine degraded images, and the DTU integrates this semantic information for object detection using a modified YOLO network. 
Our method pioneers the use of semantic data for all-weather transformations, resulting in an increase between 1.47\% to 8.80\% in mAP compared to existing methods across benchmark datasets of different weather. This highlights the potency of semantics in image enhancement and object detection, offering a comprehensive approach to improving object detection performance. Code will be available at \url{https://github.com/EnisZuo/SemOD}.
\end{abstract}




\begin{figure*}
  \centering
  \vspace{0.5em}
  \includegraphics[width=0.95\linewidth]{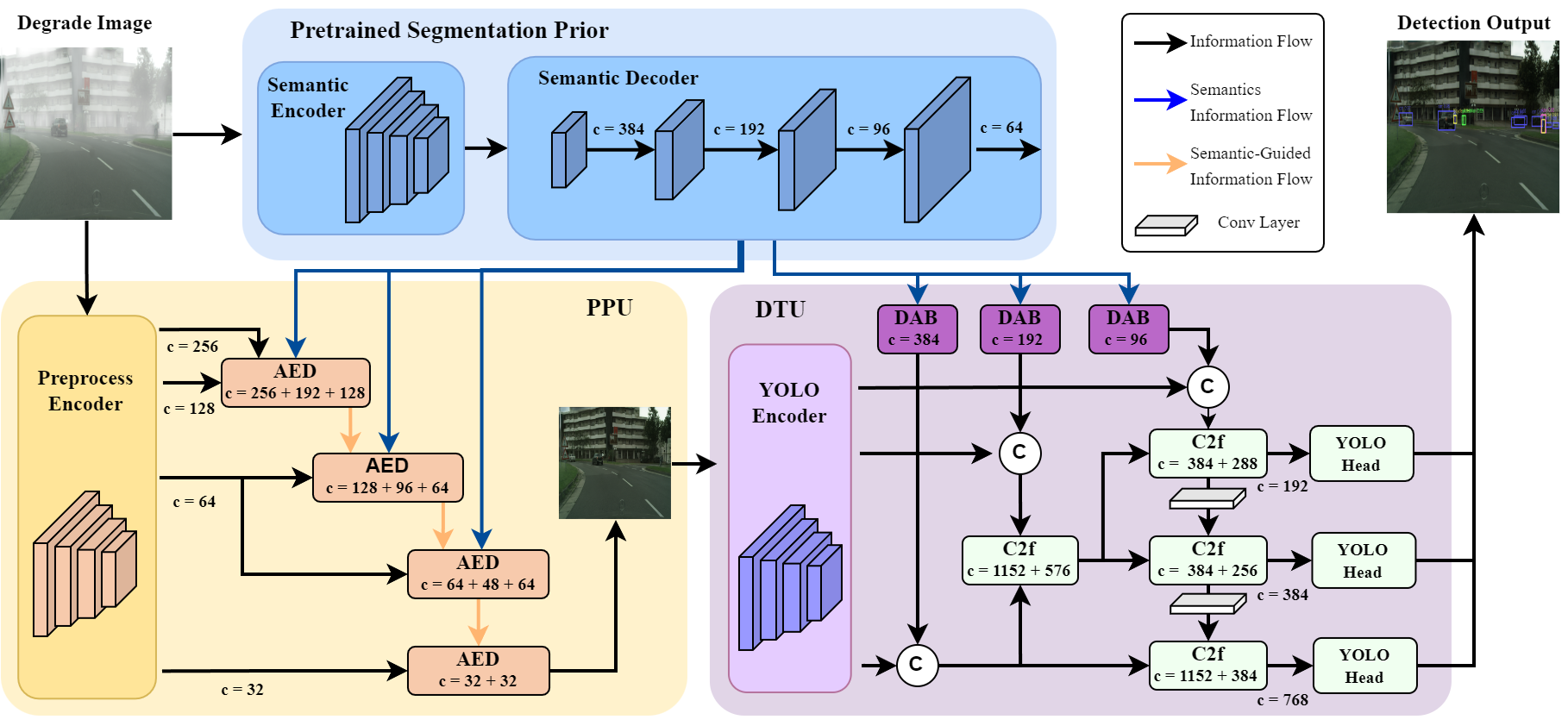}
  \caption{\textbf{Overall Model Structure} Given an input degraded image, it goes into pre-trained segmentation prior module for learning semantic information, and then our pre-processing unit(PPU) will combine both information from pre-trained segmentation prior module and PPU to generate a degraded image. The degraded image is then fed into the Object Detection Unit(DTU) for detection mission. \textcircled{c} represents feature map concatenation. ``c" stands for channel.}
  \label{fig：overall}
\end{figure*}

\section{INTRODUCTION}
\IEEEPARstart{I}{n} the burgeoning field of autonomous vehicles, camera-based perception is of paramount significance, not only for their ability to provide high-resolution spatial details, but also for the critical color information they capture\cite{vargas2021overview}. Despite significant strides in autonomous vision and tools\cite{XIA2023104120, 10192346,xiang2023v2xp}, there is a prevalent trend of models being trained and tested on datasets heavily skewed towards clear weather images\cite{fastrcnn, yolo, dosovitskiy2020image}. This bias, unfortunately, leaves them vulnerable to performance degradation under unfavorable weather conditions, such as fog, rain, or snow\cite{pei2018effects}. As the field continues to innovate, generating comprehensive models that cover a broad spectrum of computer vision tasks\cite{Shoouri2023EfficientUnderstanding}, it is imperative that the gap in performance under these severe conditions is systematically addressed, ensuring the safety and reliability of autonomous vehicles irrespective of environmental conditions, thereby paving the way for truly universal applicability.

Traditional research in the field has primarily focused on models that excel in mitigating single-domain adverse conditions such as fog, rain, or snow\cite{cai2016dehazenet, desnow, DerainRLNet2021, qian2018attentive}. While these models have been instrumental in deepening our understanding of specific weather-related challenges, their narrow focus has hindered their applicability to the broader spectrum of weather conditions that are commonly encountered in real-world driving scenarios. To address this limitation, more recent studies have pivoted towards developing models capable of handling multiple weather conditions\cite{ChenLearningModel, LiAllSearch, Valanarasu2021TransWeather:Conditions}. However, these models often prioritize weather removal performance, measured by metrics such as Peak Signal-to-Noise Ratio (PSNR), over the critical goal of object detection in autonomous vehicles. Concurrently, there has been a surge in holistic approaches aimed at improving object detection in inclement weather\cite{Wang2022R-YOLO:Weather, Ding2022CF-YOLO:Dataset, Lee2022LearningSelf-Driving}. These methods generally involve augmenting the model with additional deteriorated images or simply tweaking the detection unit to better capture objects under harsh weather conditions. Despite these advancements, these solutions frequently suffer from limitations such as a confined focus on similar weather domains (e.g., foggy and light rain conditions) or subpar performance across diverse weather conditions.


In order to tackle the dual challenges of domain adaptation and performance excellence, we proposed a \textbf{Sem}antic-enabled \textbf{O}bject \textbf{D}etection network(SemOD) for diverse weather conditions. In this network, prior knowledge of semantic segmentation provides pixel-level explanations for understanding complex environments, transforming the network from a black-box model under different weather conditions into an enhanced model structure based on semantic feature maps.
Specifically, the architecture employs a two-tiered network consisting of a Preprocessing Unit (PPU) and a Detection Unit (DTU). The PPU adopts a semantic-enhanced U-shaped net\cite{ronneberger2015u}: with its encoder deciphering the feature variances within a degraded image, it applies appropriate transformations at different scales of the feature maps to remove ambiguity or smudges based on the areas' corresponding semantic information refined in an Attention Embedded Decoder (AED). Following this, the enhanced image, along with the acquired semantic information, is connected to the downstream object detection net (DTU).
DTU incorporates a modified version of the YOLO network, adept at integrating semantic features in parallel with the original backbone output within its ``Neck" block, during which a dedicated Domain Adaptation Block (DAB) facilitates the seamless transition from the domain of semantic segmentation to object detection. This innovative orchestration of semantic information acts as an advanced attention mechanism, steering both the PPU and DTU toward enhanced performance. Notably, it delivers a superior performance increase of up to 8.80\% in mAP value over the next best contemporary methodologies.

To our knowledge, our work is the inaugural application of semantic information in all-weather image transformation and object detection. The contributions of this paper are:

1) 
This paper proposes a novel semantic enhancement framework designed for object detection under various weather conditions, utilizing semantic information to improve image quality and guide the detection process.

2) 
This paper introduces a dual-use strategy with adaptation modules, including the Attention Embedded Decoder (AED) in the Preprocessing Unit (PPU) and the Domain Adaptation Block (DAB) in the Detection Unit (DTU), to maximize the benefits of semantic module prior knowledge and significantly enhance model performance in diverse weather conditions.

3) 
This paper comprehensively evaluates the proposed model on multiple datasets and conducts a detailed study on out-of-domain datasets to demonstrate the model's adaptability to domain gaps and performance improvements.

4) This paper customized a more comprehensive dataset for verification under different weathers, and, to benefit the community, all the datasets and code are open-sourced.

\section{RELATED WORK}
\subsection{Degraded Image Transformation}
Significant strides have been made in the academic field concerning weather distortion removal from images, focusing initially on addressing individual weather phenomena such as fog, rain, and snow. Innovations include the application of Convolutional Neural Networks (CNNs) leveraging atmospheric luminosity and transmission maps for dehazing \cite{cai2016dehazenet}, and advancements in color aberration control through multiple input generation \cite{ren2018gated}. Moreover, the integration of pyramid CNNs and vision transformers \cite{zhang2018densely,song2023vision} has enriched the methodology for rain and snow removal, utilizing techniques from temporal data analysis, attention mechanisms, and advanced CNN architectures \cite{qian2018attentive,quan2019deep,you2015adherent,liu2018desnownet,li2019stacked}. Recent studies aim for a holistic approach to weather distortion removal, introducing complex modules in place of conventional convolution layers within U-Net architectures \cite{Son2020URIE:Wild}, and employing singular encoder-decoder frameworks with specialized units for minor distortions \cite{Valanarasu2021TransWeather:Conditions}. Li et al. further enhanced this approach by incorporating multiple task-specific encoders and physics-inspired tensor operations, augmented by adversarial learning \cite{LiAllSearch}. Although these methods generally perform holistic weather removal, our approach enhances the process by incorporating semantic information, thereby preserving more of the original content after removing various weather effects.

\subsection{Object Detection with Degraded Images}

In response to the critical necessity of integrating image transformation within downstream tasks for enhanced efficacy, several innovative methodologies have surfaced. One pioneering approach employs an end-to-end, deep learning-oriented framework, capable of addressing myriad weather conditions concurrently. These methodologies augment image clarity for the perception network, thereby potentiating enhanced perceptual outcomes\cite{Lee2022LearningSelf-Driving,QinDENet:Conditions}. Another end-to-end framework considers the domain adaptation in the detection and solves this problem under Foggy and Rainy Weather\cite{li2023domain}. An alternate technique progressively adapts images, originally captured under benign weather conditions, to adverse climatic scenarios. This effective interpolation bridges the chasm between two disparate domains, thereby bolstering the resilience of object detection models\cite{Wang2022R-YOLO:Weather}. Additionally, an inventive image-adaptive framework facilitates individual image enhancement for superior detection performance, proving its efficacy under both fog-laden and dimly lit conditions\cite{Liu2022Image-AdaptiveConditions}. Despite these advancements, several challenges persist: these techniques either treat image transformation and object recognition as a singular, cohesive task and train correspondingly, or they solely modify the object detectors. Consequently, despite their innovative undertones, such methods often result in a constrained focus on analogous weather domains (e.g., foggy and light rain conditions), or manifest suboptimal performance when encountering diverse domains.

\subsection{Semantic-Based Models}

Semantic segmentation, a pivotal subject in computer vision, is critical for advanced scene understanding. The advent of deep learning ushered in an era of accurate pixel-level segmentation, pioneered by Fully Convolutional Networks (FCNs) \cite{long2015fully} and U-Net \cite{ronneberger2015u}. Building on these foundations, recent advancements in large language models and transformers have further expanded the scope of vision research, leading to the development of universal segmentation networks \cite{kirillov2023segment,wang2023internimage,dosovitskiy2020image}. Following these successes, the integration of semantic prior information has been actively explored to enhance related tasks, such as image transformation and object detection. In particular, effective inpainting methods, supported by coherence priors among semantics, textures, or classes, have refined image reconstruction and contextual consistency \cite{liao2021image,tang2020local}. Through multi-scale and joint optimization strategies, a tighter synergy between image restoration and semantic segmentation has been established, enabling semantically informed refinements. In the parallel domain of video super-resolution, semantic prior-based models—most notably the GAN framework proposed by \cite{wu2019semantic}—demonstrated significant improvements by leveraging diverse texture styles across semantic classes, thereby reducing noise and recovering realistic textures through spatial feature transformation. Motivated by these insights, this paper incorporates semantic information into both the degraded pre-processing stage and the object detection stage. Such integration allows for improved restoration of meaningful content in degraded images, ultimately enhancing detection accuracy and denoising effectiveness.

\section{METHODOLOGY}
\subsection{Network Architecture}
To derive sturdy objects' bounding boxes, denoted as $\mathcal{O}$, from a visually impaired image $I$, we adopted an integrated methodology, uniting the knowledge from the spheres of image transformation and object detection. Represented in Fig. \ref{fig：overall}, initially, the impaired image $I \in \mathbb{R}^{W\times H \times3}$ was transformed into a weather-neutral image $\hat{I}\in \mathbb{R}^{W\times H \times3}$ through the Pre-Process unit (PPU), inherently enhancing visibility by eradicating visually distracting weather artifacts. Subsequently, object detection techniques were employed through Detection Unit (DTU) to excavate $\mathcal{O}$ from the resultant images.

\begin{figure}
  \centering
  \includegraphics[width=0.95\linewidth]{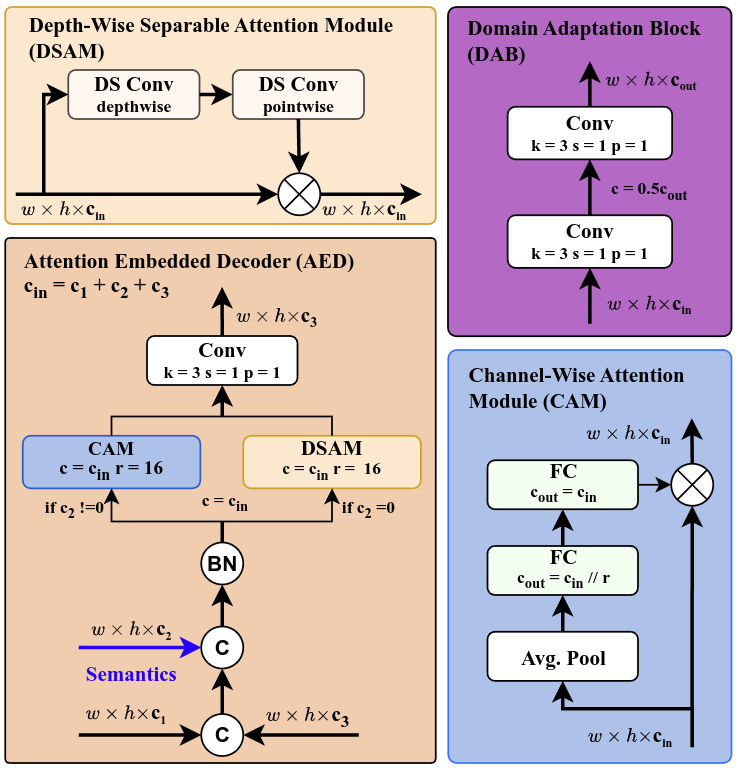}
  \caption{\textbf{Innovative Module Structure} We selectively visualize the modules that mark the key innovation within our model: attention embedded decoder(AED) in the Pre-Process Unit and domain adaptation block(DAB) in the Detection Unit. In the figure,``BN" stands for batch normalization.}
  \label{fig：part1}
\end{figure}

\subsection{Pre-Process Unit (PPU)}
\subsubsection{Structure Overview}
In the Pre-Process Unit, we aim to transform from $I$ to $\hat{I}$ under the diverse weather conditions of fog, rain, and snow. To make the transformed images $\hat{I}$ closely mimic a clear-weather counterpart of $I$, we undertook meticulous examination of images under these weather conditions. We recognized that weather effects could be dichotomized into two principal categories: visual impediments created by weather elements of varying sizes (rain, fog, snow particles) situated at diverse distances from the viewing stand and the ubiquitous obscurity and ambiguity as a result of light's inability to penetrate the particle walls. This could be encapsulated with a refined version of equations proposed in \cite{li2019heavy}:
\begin{equation}
I(x) = B(x) + \sum_i^nS_i(x)m(x) + A(1-m(x))
\end{equation}

where $x$ signifies the pixel index of the image, $I(x)$ and $B(x)$ denote the visually impaired input and the clear output respectively, \(\sum_{i=1}^{n} S_i(x)m(x)\) models the scattering effect induced by diverse particles (e.g., water droplets, dust) within the medium (e.g., fog, haze). Here, \(S_i(x)\) corresponds to the scattering effect due to the \(i\)-th particle at location \(x\), and \(m(x)\), the medium transmission map, serves as a weighting factor determining the impact of scattering on the observed intensity. $A$ characterizes the lighting conditions of this environment, and together with the latter coefficient $(1 - m(x))$ that quantifies the portion of light that is not directly transmitted but reaches the camera after atmospheric scattering, the term encapsulates the ambiguity enveloping the entire image induced by the weather. 

After the analysis of the constitution of degraded images $I\in \mathbb{R}^{W\times H \times3}$, the objective becomes generating, through pre-processing, enhanced images $\hat{I}\in \mathbb{R}^{W\times H \times3}$ that approximated the clean images $B$ as closely as possible. Many studies resorted to a U-shaped architectural framework to accomplish this, and upon scrutinizing this structure, it was evident that U-Net\cite{ronneberger2015u} performed outstandingly in removing the global atmospheric scattering effect $A(1-m(x))$ in the weather degradation model, attributable to its symmetric contracting and expanding form. Consequently, it effectively ascertained the mapping $U$ such that:
\begin{equation}
U(I(x)) = I(x) - A(1-m(x))
\end{equation}

Here, U-Net utilizes both global (via the contracting path) and local (via the expanding path) features to minimize the ambiguity $A(1-m(x))$, delivering a less noisy image $B(x) + \sum_{i}^{n} S_i(x)m(x)$. This led us to construct our Preprocess Encoder, which generates encoded feature maps at various stages- denoted as \(\Phi_i|i \in \{1,2,4,8,16\}\)- for skip connection, thus enriching the decoded feature maps for the image-wise obscureness removal using decoders.

Nonetheless, the elimination of $S_i$ from the degraded image presented a significant challenge; this was not merely a simple reconstruction task but more of an in-painting task for regions originally occluded by weather particles, where U-Net's performance fell short. This suboptimal performance arose from the characteristics of convolutions in U-Net, which primarily aggregated local and global information during reconstruction. However, these convolutions might lack sufficient context when significant portions of the image were heavily degraded (e.g., during a rainstorm) or if they concentrated on irrelevant sections of an image (e.g., focusing on road surfaces while attempting to remove snow on a car edge), preventing the model from generating novel, contextually appropriate content.

To address this issue, we incorporated semantic priors; these provided high-level contextual information, thereby enabling the model to generate plausible content for missing areas. The model then had an understanding of object boundaries and potential interactions with other items, with which a transformation could thus be applied to remove the scattering effects and mathematically expressed as follows:
\begin{equation}
B(x) = \hat{I}(x) - f(U(I(x)), S_i(x), \theta)
\end{equation}

Here, $f$ represents a stack of our attention-embedded decoders, with semantics information $\theta$- supplied by a pre-trained HRNet\cite{wang2020deep} that offers semantic feature map $\Phi_{s_i}|i\in\{2,4,8,16,32\}$- as input. HRNet is selected for its exceptional performance across various benchmark datasets. The comprehension of the general scene (via semantics) aided the model in preserving visual coherency and realism across both the filled-in and existing portions of the image. 

\subsubsection{Attention-Embedded Decoder}

Our design incorporated a decoder that accepted two feature maps $\Phi_i, \Phi_{0.5i}$ and semantic data $\theta_{0.5i}$, and in return, dispensed a decoded feature map $\Phi_{\hat{i}}|\hat{i}=0.5i$ as input for the subsequent decoder. Specifically, after rudimentary upsampling and concatenation of the feature maps to a normalized input, attention modules are instigated contingent on the presence of semantic information. Should semantic data be available, the feature maps traverse a channel-wise attention module (CAM) that imbibed the principles of squeeze-and-excitation\cite{hu2018squeeze}:
\begin{equation}
y = x\odot F_{ex}(F_{sq}(x, W_{sq}), W_{ex})
\end{equation}

This equation, as figuratively represented in Fig.2, includes squeeze and excitation layers $F_{sq}$ and $F_{ex}$ to recalibrate the original feature maps adaptively. The squeeze linear function $F_{sq}$, along with the average pooling layer, aggregates the input feature map across spatial dimensions (height and width) to produce a channel descriptor. This operation generates a global understanding of the input feature map for each channel. The excitation function $F_{ex}$ then takes the squeezed feature vector (output of $F_{sq}$); processes it through a self-gating mechanism that involves two fully-connected layers (a dimensionality-reduction layer followed by a dimensionality-increasing layer), a non-linear activation function in between, and a sigmoid activation at the end; and applies the output to the original feature map. Upon attention completion, a final convolution is invoked to reconstruct $\Phi_{\hat{i}}$ from the semantic prior-weighted feature map.

In the singular scenario where semantic information was absent and when the last decoding take place to transform $\Phi_1$ to $\hat{I}$, we deployed a strategy predicated on the depth-wise separable attention (DSAM)\cite{guo2019depthwise}, shown in Fig. \ref{fig：part1}, to capture spatial and inter-channel data for the final image output as follows:

\begin{equation}
y = x \circ \frac{1}{1 + e^{-X''}}
\end{equation}

Where $X^{''}$ is the product of two depth-wise separable convolutions with the original input $x$ and $\circ$ denotes element-wise multiplication. Hitherto, we have fashioned a decoder sequence that employs the semantic prior to guide the reconstruction process, particularly in the preliminary stages where data is scarce, yet the cascading effect is profound.

Ultimately, the efficacy of this module resides in its capacity to prioritize areas necessitating meticulous restoration and abundant guidance from the semantic maps, whilst reducing focus on regions where degradation is either uniform or negligible. This mechanism empowers the model to generate precise reconstructions, thereby considerably enhancing the overall image quality.

\subsection{Detection Unit}
\subsubsection{Structure Overview}
Subsequently after the image transformation, we extract the objects' bounding boxes $\mathcal{O}$ from the enhanced image $\hat{I}$ through non-max-suppressing $P$output of a typical YOLO\cite{yolo} detector, delineated as follows:
\begin{align}
P &= Y(\hat{I}) \\
&= {(x_i, y_i, w_i, h_i, c_{i1}, ..., c_{ic}) | i = 1, ..., K} \nonumber
\end{align}

In this equation, $P$ represents the prediction tensor outputted by the detector, $K$ illustrates the maximum number of possible bounding boxes, and $(c_{i1}...c_{ic})$ refers to the confidence scores for the $i$th bounding box  belonging to each of the $c$ classes that the model is trained to predict.

Our methodology harnessed a semantic-augmented YOLO framework to attain this $P$, predicated on the YOLO-v11\cite{khanam2024yolov11} architecture shown in Fig.1. The detection component incorporates both the refined images $\hat{I}$ and the contextually-adapted semantic data $\theta_{det}$ to yield a prediction tensor, $P=Y(\hat{I},\theta_{det}) |P\in \mathbb{R}^{B\times (4+C) \times K}$. Despite the robust pre-processing unit, the polished images may still harbor residual noise or distortions over the pristine images, denoted as $\hat{I}=B+N$. By furnishing the prediction function, $Y$, with $\theta_{det}$, the model attains heightened resilience against the noise $N$. Illustratively, discerning a roadway in $\theta_{det}$ bolsters $Y$'s confidence in detecting a car or stipulating a more precise bounding box when borders appear ambiguous by leveraging the spatial contours informed by $\theta_{det}$- even if the image $\hat{I}$ contains minor aberrations.

For realization, we incorporated the same HRNet\cite{wang2020deep} for semantic segmentation, a backbone network for feature disentanglement, a domain adapter for harmonizing semantic and detection features, and a composite neck-head network for feature orchestration and prediction articulation. During the forward pass, multi-scale features are initially extracted through the backbone, succeeded by a semantic segmentation of the input. The results of backbone and semantic segmentation outputs are amalgamated along the channel dimension to pass through the C2f layers presented in the original YOLO-v11 network \cite{terven2023comprehensive}, thereby weaving semantic and detection features into a unified canvas for efficient detection.
\subsubsection{Domain Adaptation Block}
The Domain Adaption Block (DAB) bridges the gap from the knowledge of the segmentation prior that was trained for semantic segmentation to cross-weather domain object detection. Here, the DAB conducts transformations of the semantic features to align them with the detection attributes and thus adapts the domains of semantic segmentation and object detection- it is an interim step that we leverage to ensure that the domain of semantic segmentation effectively informs and enhances the domain of object detection, regardless of the weather, for the eventual goal of robust detection.

Shown in Fig.2, upon initialization, the module creates a dual convolution that consists of a convolution2d, batch normalization, and SiLu activation. These layers were designed to acclimate the input features derived from the semantic segmentation model $\Phi_{si}\in \mathbb{R}^{\frac{W}{i}\times\frac{H}{i}\times k}$, which are intrinsically dense and pixel-specific, to the object-oriented, sparse realm of object detection $\Phi_{oi}\in \mathbb{R}^{\frac{W}{i}\times\frac{H}{i}\times k}$, thereby bolstering the prediction function $Y$.

This domain transformation process helps in consolidating the local and global context of the image, which leads to a robust and verifiably improved efficacy of the detection subsystem, and, consequently, the holistic model.

\subsection{Training}
Our model's training adheres to a sequential multi-task optimization approach, where the PPU learns to transform the degraded images first followed by the DTU acquiring the ability to yield detection from the enhanced images. In PPU, we transformed the degraded images to ($512 \times 512$) where the degraded images $I\in \mathbb{R}^{512\times 512\times 3}$ encompass all domains of adverse weather; we then employed the Charbonnier loss\cite{zhang1997parameter} as the training loss to minimize the influence of extreme outliers, as encapsulated by the equations below:
\begin{equation}
    L_{\text{{PPU}}} = \frac{1}{N} \sum_i^N {\sqrt{{(I_i - \hat{I}_i)^2 + \varepsilon^2}} - \varepsilon}
\end{equation} 
In the above equation, $I_i$ symbolizes the pixel intensity of the input image, $\hat{I}_i$ signifies the pixel intensity of the optimized image, $\varepsilon$ represents a minimal constant, and $N$ equates to the total pixel count in the image. The summation extends across all pixels within the image. 

Subsequent to the formulation of a stable $\hat{I}$, we train the detection unit employing the YOLO loss function\cite{jocher_2020}:
\begin{equation}
    L_{DTU}=\lambda_{box}L_{box}+\lambda_{class}L_{class}+\lambda_{score}L_{score}
\end{equation}
Here, $\lambda_{box}=\lambda_{class}=\lambda_{score}=1$. For the detection unit, we also enlarged the image to ($1024 \times 512$) to keep the objects' ratio uniform with the original images.

\section{Experiments}
In the following sections, we will introduce the  dataset that we use to test our experiments, experiment settings, evaluation metrics, comparison methods, quantitative results and qualitative results.
\subsection{Dataset}
\noindent\textbf{Cityscapes Dataset.} In our quest for robust object detection under challenging weather conditions, we turned to the Cityscapes~\cite{Cordts2016Cityscapes} dataset, which is rich in its diversity across various climatic scenarios. From this collection, we sourced:
\begin{itemize}
    \item 3,475 pristine (sunny) images\cite{Cordts2016Cityscapes}.
    \item 10,425 foggy captures, with visibilities of 150, 300, and 600 meters, courtesy of~\cite{sakaridis2018semantic}. \textcolor{black}{Foggy Cityscapes is established by simulating the fog of different intensity levels on the Cityscapes images, which generates the simulated three levels of fog based on the depth map and a physical model.} 
    \item 1,062 rainy images, which involve 36 variations of rain intensity on 295 selected images, as provided by~\cite{Hu_2019_CVPR}.
\end{itemize}

\textcolor{black}{Such datasets are widely adopted benchmarks in the community, providing standardized and reproducible evaluation protocols for fair comparison across different methods. }

\noindent\textbf{Customized dataset.}
To achieve more comprehensive coverage of weather conditions, we performed data augmentation and data generation on existing datasets to enrich and create more diverse datasets under various weather conditions. Following Transweather\cite{Valanarasu2021TransWeather:Conditions}, we incorporated:
\begin{itemize}
    \item The RainDrop dataset, which consists of 1,069 images\cite{qian2018attentive}.
    \item A subset of Snow100K\cite{liu2018desnownet}, from which we selected 13,283 images to represent snowy conditions.
\end{itemize}

\noindent\textbf{Dataset classes.}
For the purpose of bounding box extraction across all these datasets, our primary focus was on core traffic participants. Our detection classes consisted of car, pedestrian, truck, bus, rider, bicycle, and motorcycle. To facilitate the extraction process, we employed \cite{cityscapetococo} which enabled us to attain 2D bounding boxes for the Cityscape datasets efficiently. Meanwhile, for the Snow100K dataset, annotations were done manually. Additionally, we integrated clear weather images from Cityscape into the enhanced datasets to act as a benchmark for detection under other weather conditions. Access links for the unified dataset and annotations can be found in our Github repository: \url{https://github.com/EnisZuo/SemOD}

\begin{table}[h]
\centering
\vspace{-0.5em}
\caption{\textcolor{black}{\textbf{Pre-Process Unit.} Our model's performance exceeds all state-of-the-art methods in the domain of multi-weather transformation. The efficacy of this improvement is further demonstrated in object detection tasks.}}
\vspace{-0.5em}
\label{table:pre-process}
\begin{tabular}{p{3.2cm}p{1.4cm}p{1.4cm}}
\toprule
Methods  & PSNR $\uparrow$& SSIM $\uparrow $\\
\addlinespace[-0.3em]
\midrule
Swin-IR (2021)\cite{liang2021swinir}         &  23.23  & 0.869  \\
Urie (2020)\cite{Son2020URIE:Wild}         &  24.96   & 0.863   \\
All-in-One (2021)\cite{LiAllSearch}      &  24.71 & 0.898 \\
Transweather (2022)\cite{Valanarasu2021TransWeather:Conditions}         & 27.74 & 0.912 \\
\textbf{SemOD(Ours)} & \textbf{29.41} & \textbf{0.924} \\
\bottomrule
\end{tabular}
\vspace{-0.5 em}
\end{table}

\begin{table*}[h]
\centering
\vspace{0.5em}
\caption{\textbf{Ablation Study}. \textcolor{black}{Through an analysis of four distinct model setups, we systematically removed each block to demonstrate its individual contribution to the overarching performance.}}
\label{table:ablation}
\vspace{-0.5em}
\begin{tabular}{>{\raggedright\arraybackslash}p{2.5cm}|p{1.1cm}p{0.6cm}p{0.7cm}|p{1.1cm}p{0.6cm}p{0.7cm}|p{1.1cm}p{0.6cm}p{0.7cm}|p{1.1cm}p{0.6cm}p{0.7cm}}
\toprule
 \textbf{Weather} & \multicolumn{3}{c|}{\textbf{Foggy}} & \multicolumn{3}{c|}{\textbf{Rainy}} & \multicolumn{3}{c|}{\textbf{Snowy}} & \multicolumn{3}{c}{\textbf{Sunny}}\\
\textbf{Methods}& $mAP_{50-95}$ & $mAP_{50}$&$mAP_{75}$& $mAP_{50-95}$ & $mAP_{50}$ & $mAP_{75}$& $mAP_{50-95}$ & $mAP_{50}$ & $mAP_{75}$& $mAP_{50-95}$ & $mAP_{50}$ & $mAP_{75}$ \\
\midrule
\textcolor{black}{Yolo-v11} & 27.09 & 31.53 & 28.15 & 26.11 & 44.53 & 25.77 & 17.33 & 26.92 & 18.98 & 26.80 & 44.68 & 27.97\\
\textcolor{black}{PPU+Yolo-v11} & 31.51 & 47.38 & 34.35 & 31.34 & 46.37 & 34.03 & 17.00 & 27.47 & 18.05 & 33.08 & 49.71 & 35.95\\
\textcolor{black}{PPU+Yolo-v11+Sem} & 34.50 & 50.87 & 37.32 & 31.68 & 46.87 & 33.55 & 22.72 & 35.10 & 24.56 & 33.97 & 51.20 & 36.78\\
\textcolor{black}{SemOD} & \textbf{36.16} & \textbf{51.95} & \textbf{40.06} & \textbf{32.61} & \textbf{48.13} & \textbf{36.73} & \textbf{27.29} & \textbf{37.43} & \textbf{30.09} & \textbf{35.75} & \textbf{52.80} & \textbf{38.76} \\
\bottomrule
\end{tabular}
\vspace{-1em}
\end{table*}

\subsection{Experiment settings}
We merged data of the same weather conditions from the datasets mentioned above, and the training dataset was shuffled randomly while preserving the individual test sets from each dataset. To assess the performance of models trained universally under each distinct weather scenario more fairly, we split the training and validation sets with a ratio of 4:1.
we resized each sample in the dataset to a size of $512\times512$ as the input for the Pre-Process Unit (PPU). The transformed images from PPU are then resized to $512\times1024$ for the detection unit (DTU) to output detection bounding boxes- a resize that keeps the object dimensions consistent with the original images. The same resizing and similar image flow are applied to other benchmark models in order to conduct fair comparison.

\textcolor{black}{Every model, including the SOTA methods used for comparison, was trained and assessed on a single Nvidia RTX 3090 GPU with the learning rate 0.0005. The training was conducted with a train batch size of 12 and a test batch size of 16, using SGD as the optimizer and a weight decay of 0.0001. Training of all models in both processing steps commences with an initial duration of 50 epochs.} Adhering to the manners used by Transweather~\cite{Valanarasu2021TransWeather:Conditions} and Yolo-v11~\cite{khanam2024yolov11}, we report metric values on validation sets, where higher values signify superior performance.

\subsection{Evaluation Metrics and Comparison Methods}
We evaluate detection quality with COCO-style mean Average Precision \cite{lin2014microsoft}, reporting $mAP_{50}$ (AP at IoU$=0.50$), $mAP_{75}$ (AP at IoU$=0.75$), and $mAP_{50\text{--}95}$, the mean AP averaged over IoU thresholds $\{0.50,0.55,\ldots,0.95\}$. Our comparisons use YOLOv11 as detection baselines, and it cover weather-removal–plus–detector pipelines (DENet~\cite{qin2022denet}, UEMYolo~\cite{lee2022learning}, Urie+Yolo \cite{Son2020URIE:Wild}, and TransWeather+Yolo \cite{Valanarasu2021TransWeather:Conditions}), and domain-adaptive detectors (DA-Faster~\cite{chen2018domain}, UaDAN~\cite{guan2021uncertainty}, and DA-detect~\cite{li2023domain}); for fairness, all methods use the same resizing scheme and validation splits. 



\begin{table*}[h]
\centering
\caption{\textbf{Entire Model}. 
\textcolor{black}{Comparison across four weather conditions using COCO-style mAP: $mAP_{50}$ (AP@0.50), $mAP_{75}$ (AP@0.75), and $mAP_{50\text{--}95}$ (mean AP over IoU thresholds $0.50{:}0.05{:}0.95$). All values are percentages from a single trained model evaluated on weather-specific validation sets under an identical resizing and inference pipeline. Baselines include weather-removal–plus–detector pipelines (DENet~\cite{qin2022denet}, UEMYolo~\cite{lee2022learning}, Urie+Yolo \cite{Son2020URIE:Wild}, TransWeather+Yolo \cite{Valanarasu2021TransWeather:Conditions}) and domain-adaptive detectors (DA-Faster~\cite{chen2018domain}, UaDAN~\cite{guan2021uncertainty}, DA-detect~\cite{li2023domain}); detection is performed with YOLOv11 in our pipeline for fair comparison. SemOD (Ours) attains the best results across all metrics and we report in the \emph{Improvement} row its absolute margin over the strongest competing method in each column.
}}
\label{table:entire}
\vspace{-0.5 em}
\begin{tabular}{>{\raggedright\arraybackslash}p{2.5cm}|p{1.1cm}p{0.6cm}p{0.7cm}|p{1.1cm}p{0.6cm}p{0.7cm}|p{1.1cm}p{0.6cm}p{0.7cm}|p{1.1cm}p{0.6cm}p{0.7cm}}
\toprule
 \textbf{Weather} & \multicolumn{3}{c|}{\textbf{Foggy}} & \multicolumn{3}{c|}{\textbf{Rainy}} & \multicolumn{3}{c|}{\textbf{Snowy}} & \multicolumn{3}{c}{\textbf{Sunny}}\\
\textbf{Methods}& $mAP_{50-95}$& $mAP_{50}$&$mAP_{75}$& $mAP_{50-95}$ & $mAP_{50}$ & $mAP_{75}$& $mAP_{50-95}$ & $mAP_{50}$ & $mAP_{75}$& $mAP_{50-95}$ & $mAP_{50}$ & $mAP_{75}$\\
\midrule
 DENet & 24.45 & 41.39 & 24.85 & 21.34 & 37.58 & 20.57 & 14.12 & 25.17 & 13.87 & 24.67 & 42.74 & 24.72 \\
 UEMYolo & 26.64 & 43.59 & 27.57  & 25.67 & 43.44 & 25.24 & 14.18 & 23.67 & 15.28 & 26.35 & 43.59 & 27.39 \\
 Urie + Yolo & 29.87 & 44.39 & 32.38 &  29.07 & 43.14 & 31.30  & 16.66 & 27.09 & 17.17  & 32.32 & 48.73 & 34.69 \\
 \textcolor{black}{UaDAN} & 25.29 & 41.10 & 27.35 & 21.26 & 36.81 & 22.92 & 15.02 & 25.37 & 16.04 & 25.53 & 42.63 & 27.43 \\
 \textcolor{black}{DA-Faster} & 25.19 & 41.00 & 27.25 & 21.00 & 36.54 & 22.66 & 15.09 & 25.44 & 16.10 & 25.50 & 42.59 & 27.39 \\
  \textcolor{black}{DA-detect} & 26.49 & 42.30 & 28.55 & 26.46 & 44.00 & 28.12 & 17.20 & 24.55 & 15.22 & 26.00 & 43.09 & 27.89 \\
   Transweather + Yolo & 30.80 & 45.65 & 33.34  & 30.15 &  44.28& 33.30  & 17.07 &  27.71 & 18.00 & 33.34 & 50.04 & 36.33 \\
 \addlinespace[2pt]
 SemOD (Ours) & \textbf{36.16} & \textbf{51.95} & \textbf{40.06} & \textbf{32.61} & \textbf{48.13} & \textbf{36.73} & \textbf{27.29} & \textbf{37.43} & \textbf{30.09} & \textbf{35.75} & \textbf{52.80} & \textbf{38.76} \\
 \midrule
 Improvement & 4.75 & 5.03 & 5.89 &  1.91 & 2.67 & 2.67 & 9.76 & 8.80 & 11.47 & 1.81 & 1.47 & 1.63 \\
\bottomrule
\end{tabular}
\vspace{-2em}
\end{table*}

\subsection{Quantitative Results}
\textbf{\textcolor{black}{Pre-Process Unit Analysis.}}
\textcolor{black}{In gauging the efficacy of the Pre-Process Unit, we meticulously adhered to the benchmarks set by \cite{Valanarasu2021TransWeather:Conditions}, deploying two prominent metrics: PSNR (Peak Signal-to-Noise Ratio) and SSIM (Structural Similarity Index Measure). PSNR quantifies the fidelity discrepancy between an original image and its modified counterpart, with superior fidelity indicated by a heightened PSNR. Conversely, SSIM assesses variances in structural nuances, luminosity, and texture, delivering a holistic, perceptually salient evaluation. Its values span from -1 to 1, with a perfect score of 1 denoting identical imagery.}

\textcolor{black}{Table. \ref{table:pre-process} illustrates that our module surpasses contemporary pinnacle models, registering an elevation of at least 6.02\% in PSNR and 1.32\% in SSIM. This underscores the mastery of our semantic-enabled reconstruction. Additionally, beyond mere metric enhancement, our Pre-Process Unit adeptly emphasizes regions deemed pivotal by semantic information, which is further elucidated by subsequent object detection metrics and qualitative analyses.}

\noindent \textbf{Ablation Study.} 
To study the contribution of each module in achieving such an object detection performance, we started from the plain Yolo-v11 network and added components one by one and thus identified four structures: \textcolor{black}{(1)Yolo-v11 detection module (2)PPU + Yolo-v11, where PPU denote pre-process unit (3) PPU + Yolo-v11 + Semantic module (Ours w/o domain adaption module) (4)SemOD (PPU + Yolo-v11 + Semantic module + DAB). }
All experiments are conducted with the same training parameters on 4 cross-domain datasets. The ablation study in Table. \ref{table:ablation} clearly justifies the positive effect of each proposed module across all datasets.

Our Pre-Process Unit (PPU), mirroring other image transformation components, resulted a pronounced uplift in object detection efficacy. Subsequent incorporation of semantic information markedly augmented the robustness of detection, especially in Snowy conditions where a domain gap exists. A simple domain adaptation block further accentuated this enhancement, solidifying an unmatched performance across datasets. Such outcomes attest that our module deployment is not merely theoretically cogent but also pragmatically pivotal in advancing the paradigm of semantic-enabled object detection.

\begin{figure*}[h]
\centering
\begin{subfigure}{.30\textwidth}
\centering
  \includegraphics[width=1.0 \linewidth]{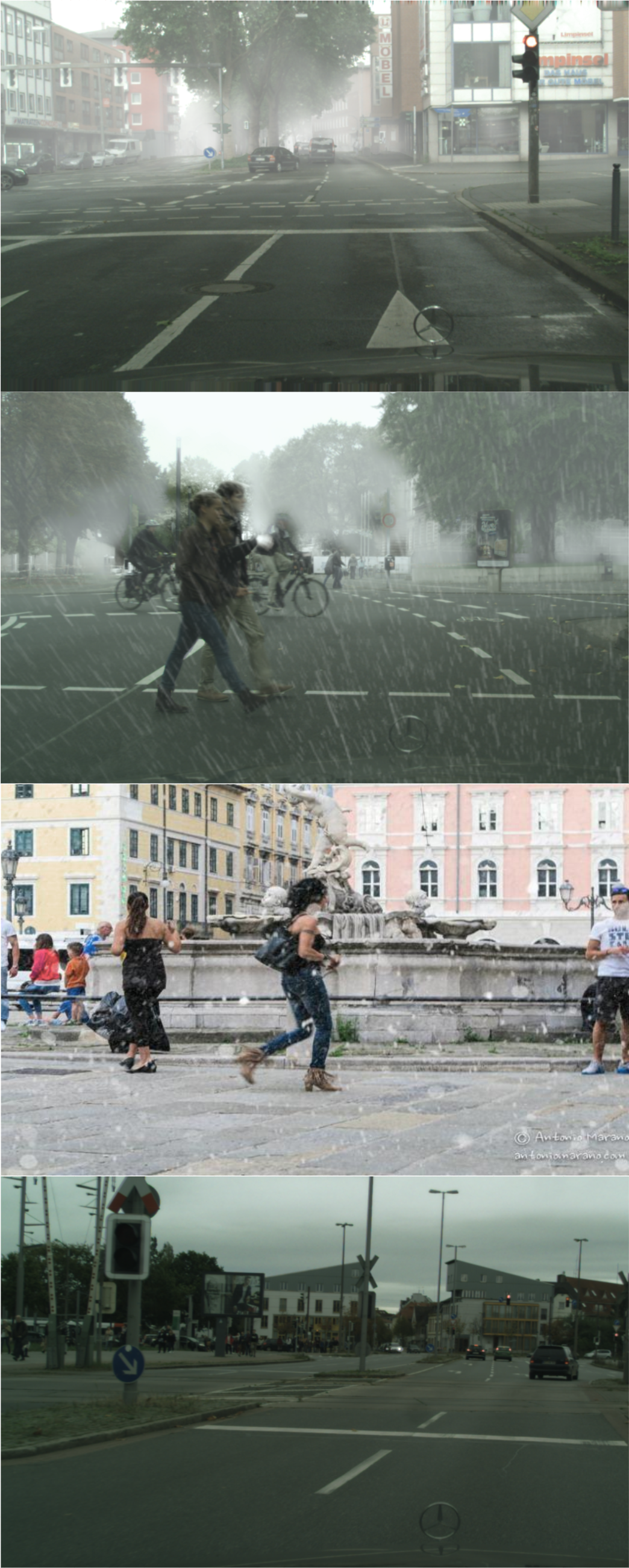}  
  \caption{}
  \label{fig:v_ori_foggy}
\end{subfigure}
\begin{subfigure}{.30\textwidth}
\centering
  \includegraphics[width=1.0 \linewidth]{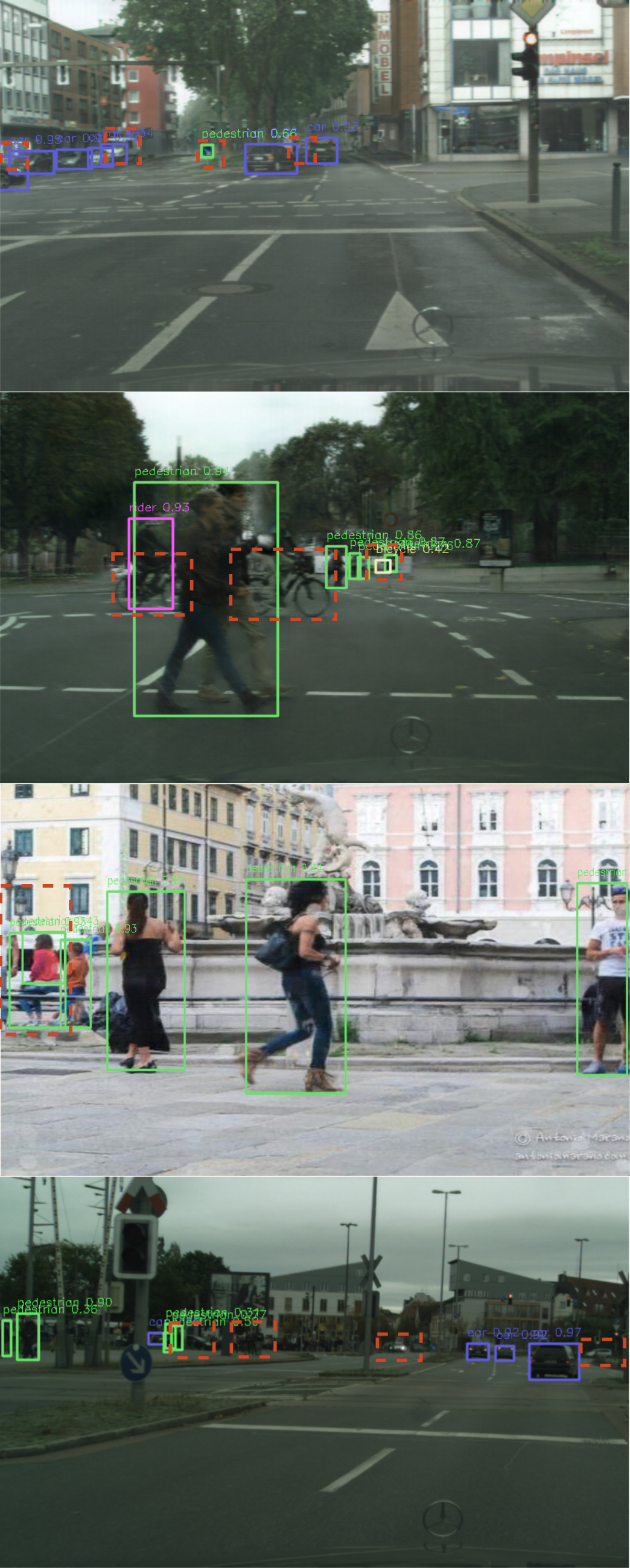}  
    \caption{}
  \label{fig:ty-foggy}
\end{subfigure}
\begin{subfigure}{.30\textwidth}
\centering
  \includegraphics[width=1.0 \linewidth]{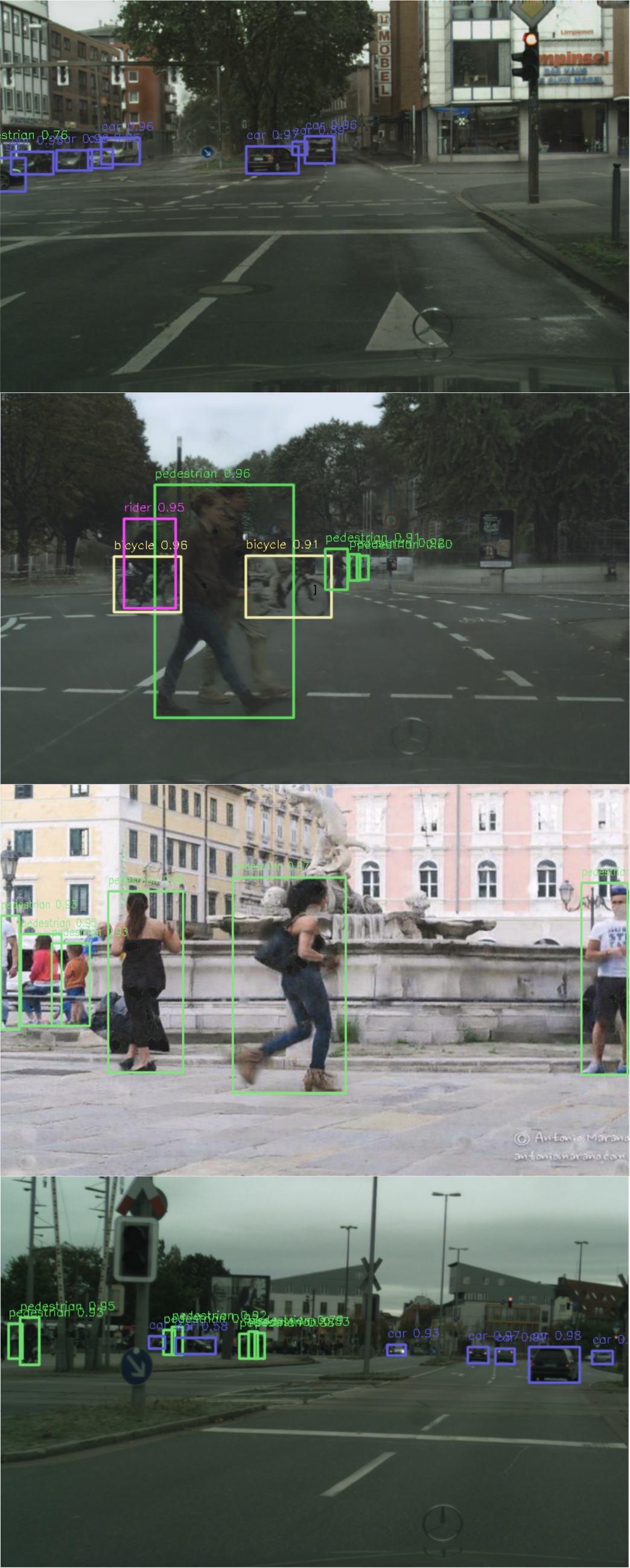}  
 \caption{}
  \label{fig:illus_US_c}
\end{subfigure}
	\caption{Qualitative visualization results are presented for the validation set of four distinct weather datasets. The first row depicts conditions during \textbf{foggy} weather, the second row shows \textbf{rainy} weather, the third row illustrates \textbf{snowy} weather, and the final row portrays \textbf{sunny} weather. (a) Original dataset images. (b) Suboptimal model(Transweather+Yolo$_{v11}$) output results. (c) Our method's results. Note that both(b)(c) include weather removal effect and detection results. All the solid line bounding boxes are the final detection results and different colors represent different classes. The highlighted dashed red bounding boxes are the wrong detection results compared with (c). }
	\label{fig:visualization}
\end{figure*} 

\noindent\textbf{Entire Model.} The performance comparison of different models under various weather scenarios can be found in Table \ref{table:entire}. In order to comprehensively assess the performance of SemOD, we compared it with two integrated solutions specifically designed for object detection under adverse weather conditions based on the YOLO framework\cite{qin2022denet,lee2022learning}. Additionally, to evaluate our model's performance in weather removal, we compared it against the current two best-performing image transformation methods\cite{Valanarasu2021TransWeather:Conditions,LiAllSearch}, and integrated them with YOLO-v11 for comparison across four different weather conditions.

As shown in Table \ref{table:entire}, our method demonstrates significant improvements in mAP compared to the suboptimal methods across all adverse weather conditions, with increases of 5.03\% for foggy, 2.67\% for rainy, and 8.8\% for snowy scenarios, respectively. It is noteworthy that the improvements are even more pronounced on the customized snowy weather dataset that is not based on Cityscapes, due to the larger and more distinct domain differences between this dataset and the Cityscapes-based dataset. Supported by our semantic module, our model not only exhibits the best enhancement performance but also reflects the greater ability of this approach to substantially reduce domain divergence effects.
Furthermore, it is noteworthy that even in clear weather conditions without adverse weather interference, our model outperforms the YOLO-v11 detection model (with a 1.47\% improvement). This result indicates that with the enhanced support of the semantic module, the accuracy of detection has also been enhanced. Therefore, through quantitative comparisons across different weather datasets, the superiority of our approach is evident.

\begin{table}[t]
\centering
\small
\caption{Estimated per-frame inference time (ms) across multiple components.}
\label{tab:runtime_estimates}
\begin{tabular}{l c}
\toprule
\textbf{Algorithm Component} & \textbf{Time (ms/frame)} \\
\midrule
YOLOv11                  & 6--12  \\
PPU + YOLOv11            & 11--23 \\
PPU + YOLOv11 + Sem      & 22--43 \\
SemOD (with DAB)         & 23--46 \\
\bottomrule
\end{tabular}
\vspace{-3mm}
\end{table}

\noindent\textbf{Inference time.} \textcolor{black}{To address real-time applicability and computational cost, we report estimated end-to-end per-frame latency under the same setting as our accuracy experiments (single NVIDIA RTX~3090, batch=1; PPU input \(512\times512\), detector input \(512\times1024\)). As summarized in Table~\ref{tab:runtime_estimates}, relative to the plain detector, SemOD adds only \(\sim17\text{--}34\,\mathrm{ms}\) per frame while delivering the reported accuracy gains; the DAB alignment accounts for merely \(\sim1\text{--}3\,\mathrm{ms}\) of the total latency. These results indicate that our method is deployable in real time on commodity GPUs.}

\subsection{Qualitative Results}

\setlength{\belowcaptionskip}{-8pt}


In the qualitative assessment, we set our model against its next-best alternative, \textcolor{black}{``Transweather + Yolo-v11"}, across four distinct weather scenarios, as depicted in Fig. \ref{fig:visualization}. 
We not only compared the detection performance of the models but also assessed the effects after weather removal. By comparing the first rows of (b) and (c) at an enlarged scale, we observed that our approach achieves a higher level of scene restoration after weather removal. Particularly, when comparing images, such as roadside billboards and text, we found that the clarity and sharpness were notably improved.
Inspecting the bounding boxes of all classes, it's evident that our model, SemOD, consistently delivers higher confidence, superior accuracy, and fewer false positives. Specifically, SemOD's superiority extends as objects reside farther away from where the images are taken. In fact, SemOD not only rectifies inaccurate or even false detection boxes but also captures several diminutive objects ignored by the alternative, spanning from pedestrians to bicycles to vehicles. For example, looking at the image under the sunny weather scenario, where adverse weather is no longer a distractor, our model can detect some objects that are far away. This observation is a testament to our theoretical analysis that without contextual information provided by semantics prior, traditional models are inferior in generating logical and meaningful content to replace weather effects and informed bounding boxes in areas heavily degraded by different weathers.

Additionally, this qualitative analysis corroborates our understanding that SemOD, incorporating semantics, is more robust towards domain gaps among different datasets: Sunny, foggy, and rainy datasets are all generated images from the Cityscape dataset. Snowy weather images are, on the other hand, selected from the Snow100K dataset and thus have different lighting, architecture, and traffic patterns, as shown in our visualization. Here, SemOD yields clearer weather removal images and bounding boxes with higher confidence scores compared to the next best alternative. 
However, other models sacrificed a certain level of environmental interpretability when dealing with different weather conditions, resulting in detection outcomes that did not meet our expectations, especially when there was a significant domain gap, which was more evident in such cases.
SemOD attained common and crucial features under adverse weather images through expanded interpretation ability offered by the semantic network.

\section{Conclusion}
In this study, we introduced ``SemOD", a semantic-enhanced object detection network tailored for robust performance in various weather conditions, including fog, rain, snow, and clear skies. Our network comprises a preprocessing unit and a detection unit. We not only elucidated the amplified benefits of semantic information at two critical model stages—image transformation and object detection—but also rigorously substantiated this synergy through extensive experimentation. This integration significantly enhances the mean Average Precision of object detection, surpassing the state-of-the-art (SOTA) across all comparisons, with improvements ranging from 1.47\% in clear skies to 8.80 \% in snowy conditions.



\section*{REFERENCES}

\def\refname{\vadjust{\vspace*{-1em}}} 

\bibliographystyle{IEEEtran}
\bibliography{reference}

\begin{IEEEbiography}[{\includegraphics[width=1in,height=1.25in,clip,keepaspectratio]{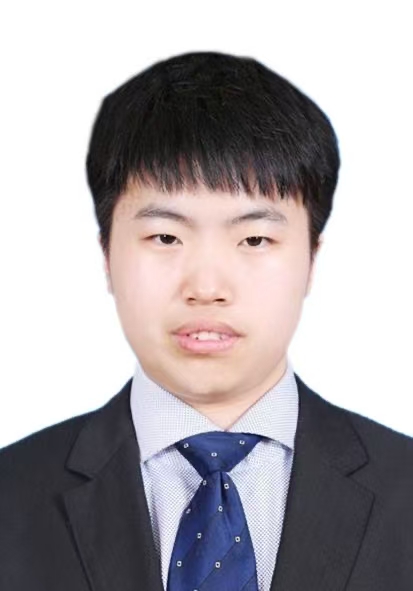}}]{Aiyinsi Zuo } received dual B.S. degree in Electrical and Computer Engineering and Business: Entrepreneurship from the University of Rochester. During his undergraduate studies, he contributed to research projects in spatiotemporal forecasting, semantic-enabled object detection, and real-time perception for autonomous driving. He also served as a teaching assistant, supporting courses in Mechatronics, Embedded Systems, and Operations Strategy. Currently, he is pursuing his Ph.D. at CREOL, the College of Optics and Photonics, University of Central Florida, where he focuses on photonics research. He is also a dedicated student member of SPIE.
\end{IEEEbiography}

\begin{IEEEbiography}[{\includegraphics[width=1in,height=1.25in,clip,keepaspectratio]{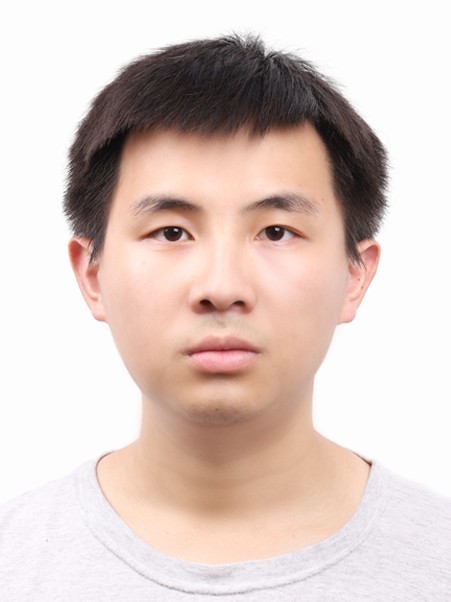}}]{Zhaoliang Zheng } received a B.E. degree in Process Equipment and Control Engineering from Dalian University of Technology, China, in 2017, and an M.S. degree in Mechanical and Aerospace Engineering from the University of California, San Diego, CA, USA, in 2019. He is currently pursuing a Ph.D. degree in Electrical and Computer Engineering at the UCLA Mobility Lab, University of California, Los Angeles, CA, USA. His research interests include robotics and embedded systems, sensing and planning for autonomous driving, and V2X cooperative perception.
\end{IEEEbiography}




\end{document}